\documentclass[a4paper]{article}

\usepackage{INTERSPEECH2022}

\usepackage{url}
\usepackage{hyperref}
\usepackage[capitalize,noabbrev]{cleveref}
\usepackage{xcolor}

\usepackage[normalem]{ulem}

\def\footurl#1{\footnote{\url{#1}}}

\title{Short-Term Word-Learning in a Dynamically Changing Environment}

\name{Christian Huber$^1$, Rishu Kumar$^2$, Ond\v{r}ej Bojar$^2$ and Alexander Waibel$^{1,3}$}

\address{
  $^1$Interactive Systems Lab, Karlsruhe Institute of Technology, Karlsruhe, Germany\\
  $^2$Charles University, Faculty of Mathematics and Physics,\\ Institute of Formal and Applied Linguistics, Prague, Czech Republic\\
  $^3$Carnegie Mellon University, Pittsburgh PA, USA}
\email{christian.huber@kit.edu, lastname@ufal.mff.cuni.cz, alexander.waibel@cmu.edu}

\begin{document}

\maketitle
\begin{abstract}
Neural sequence-to-sequence automatic speech recognition (ASR) systems are in principle open vocabulary systems, when using appropriate modeling units. In practice, however, they often fail to recognize words not seen during training, e.g., named entities, numbers or technical terms.
To alleviate this problem, \cite{huber2021instant} proposed to supplement an end-to-end ASR system with a word/phrase memory and a mechanism to access this memory to recognize the words and phrases correctly.
In this paper we study, a) methods to acquire important words for this memory dynamically and, b) 
the trade-off between improvement in recognition accuracy of new words and the potential danger of false alarms for those added words. 
We demonstrate significant improvements in the detection rate of new words with only a minor increase in false alarms (F1 score 0.30 $\rightarrow$ 0.80), when using an appropriate
number
of new words.
In addition, we show that important keywords can be extracted from supporting documents and used effectively. 
\end{abstract}
\noindent\textbf{Index Terms}: speech recognition, one-shot learning, new-word learning

\section{Introduction}

Neural sequence-to-sequence systems deliver state-of-the-art performance for automatic speech recognition (ASR). When using appropriate modeling units, e.g., byte-pair encoded characters, these systems are in principle open vocabulary systems. In practice, however, they often fail to recognize words not seen during training, e.g., named entities, numbers or technical terms.

To alleviate this problem, \cite{huber2021instant} proposed to supplement an end-to-end ASR system with a word/phrase memory and a mechanism to access this memory to recognize the words and phrases correctly.
After the training of the ASR system, and when it has already been deployed, a relevant word can be added or subtracted instantly without the need for further training.


This is achieved by, 
a) a memory-attention layer which
predicts the availability and location of relevant information in the memory,
and b) a memory-entry-attention layer which extracts the information of a memory entry.

In this paper we study, 
a) methods to acquire specialized words for this memory and, b) 
the trade-off between improvement in recognition accuracy of new words and the potential danger of false alarms for those added words. Therefore, we extensively evaluate this system in an online low-latency setup.

The ASR model described above outputs uncased text without punctuation.
Therefore, we run a casing and punctuation model afterwards which reconstructs the casing of each word and inserts the punctuation. This model consists of a transformer encoder \cite{vaswani2017attention,pham2019very} which is run after the beam-search and outputs for each word if the word should be uppercased and if any punctuation should be emitted after the word or not.

To use the model in an online low-latency setup \cite{nguyen2021superhuman}, we do the following: The model waits for a chunk of acoustic frames with at least a predetermined duration to arrive. Then beam search is run with this input chunk. The beams (called unstable hypotheses) are then given to a stability detection 
component 
which returns a stable hypothesis,
e.g. the common prefix of all hypotheses. After that, the part of audio corresponding to the stable hypothesis is cut out (via an alignment) and the model waits for more audio frames.

\section{Experiments and Results}

We extended the model proposed in \cite{huber2021instant} with the ability to correctly do the casing of the new words supplied through the new words list. This is done by adapting the casing and punctuation model by using internals of the ASR model, namely the attention over the memory entries. For each word the beam-search has outputted, all the predicted memory entries (from the memory-attention layers) are compared with the word. If there is a match, the casing from the new words list is taken. Therefore, whenever a new word is recognized by the model, the correct casing from the new words list is outputted.

Furthermore, we evaluate the system in two different scenarios: First, when an operator adds new words to the system, and second, when new words are extracted from other sources, e.g. slides.

\subsection{Data}

For the first scenario, we use eight talks from the ELITR testset \cite{sltev:eacl:2021} with a total length of 3.7 hours.
For the second scenario, we use ten talks from the EMNLP 2020 conference\footnote{\url{https://2020.emnlp.org}} with a total length of 1.6 hours. 
Along with the EMNLP talks, the papers and the slides of the talks are available. The text from the papers (excluding the references) is extracted with pdftotext\footnote{\url{https://en.wikipedia.org/wiki/Pdftotext}}, the text from the slides is extracted with Tesseract\footnote{\url{https://github.com/tesseract-ocr/tesseract}}, since we only had access to screenshots of the slides. We also cleaned the transcripts of all talks from typos, so that they could serve as a reliable reference.

\begin{figure}[t]
    \centering
    \includegraphics[trim={0 0 14cm 0},clip,width=0.9\linewidth]{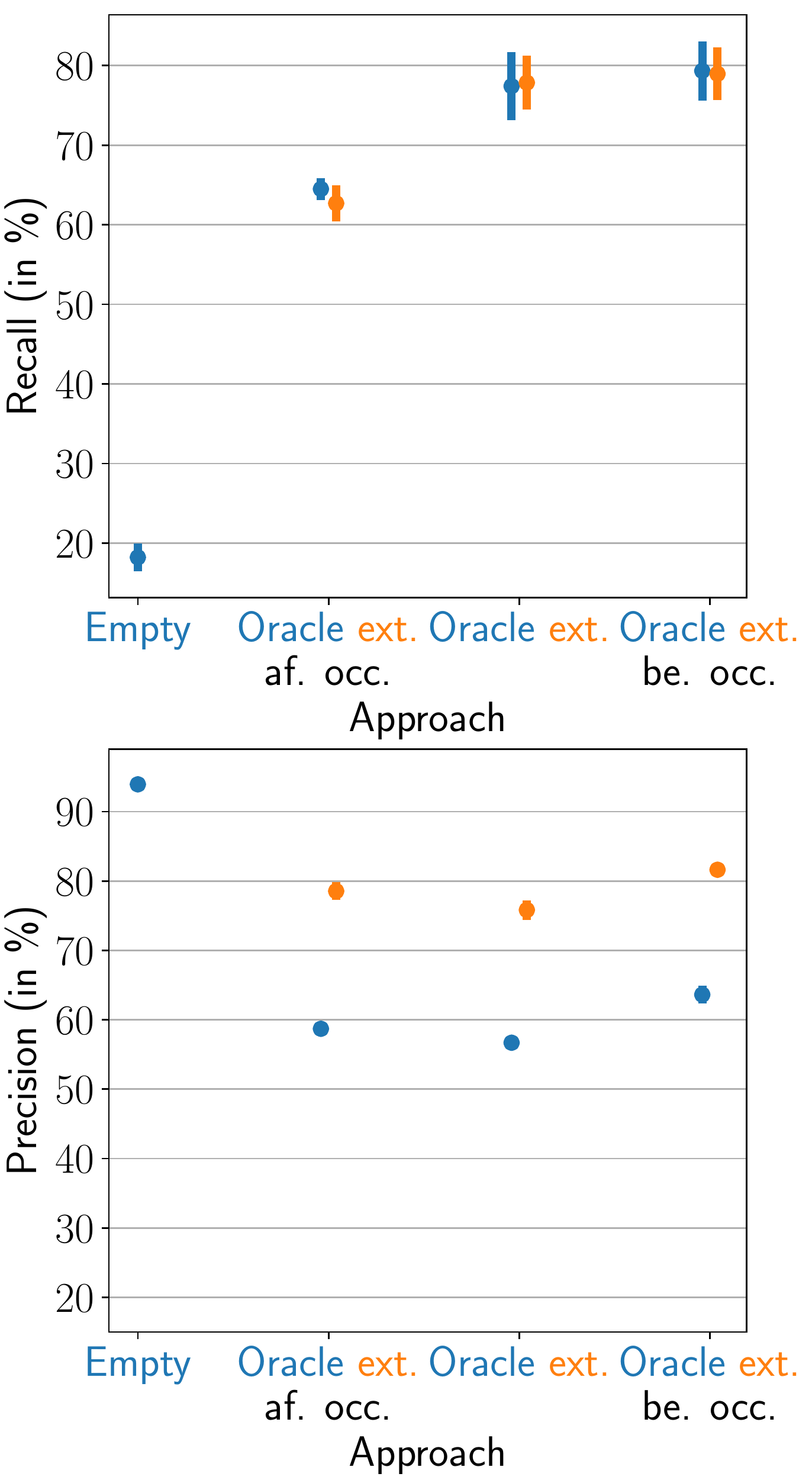}
    \caption{Recall and precision for the evaluation of the memory ASR-worker on the ELITR testset.}
    \label{fig:testsetELITR}
\end{figure}

\begin{figure}[t]
    \centering
    \includegraphics[trim={14cm 0 0 0},clip,width=0.9\linewidth]{images/plot_ELITR.pdf}
    \caption{mWER and F1-score for the evaluation of the memory ASR-worker on the ELITR testset.}
    \label{fig:testsetELITR2}
\end{figure}

\subsection{Extraction of the New Words List}

We tried different methods to extract a list of new words from the document, and we ended up taking all the words of the document which are not in the training data of the ASR model. This method is simple and performed well when looking at the output words\footnote{Furthermore this method is very effective in finding errors in the transcript.}. For example in the talk \texttt{rehm\_long},\footurl{https://github.com/ELITR/elitr-testset/tree/master/documents/rehm-language-technologies} we have extracted the following list of new words:
pipelining, Friem, iAnnotate, MQM, LSPs, eServices, semantification, Aljoscha, Cortana, workflows, DFKI, annotating, NLP.

With this method, we extracted 134 terms from the ELITR testset
reference transcripts, 
148 from the EMNLP testset
reference transcripts, 
865 from the EMNLP papers and 584 from the EMNLP slides.
Note that, in contrast to the new-words testset in \cite{huber2021instant}, the list of new words is created automatically and might differ from a list of new or rare words a human might select.

\subsection{Evaluation with the Help of an Operator}

\begin{figure}[t]
    \centering
    \includegraphics[trim={0.15cm 0 17.9cm 0},clip,width=1.0\linewidth]{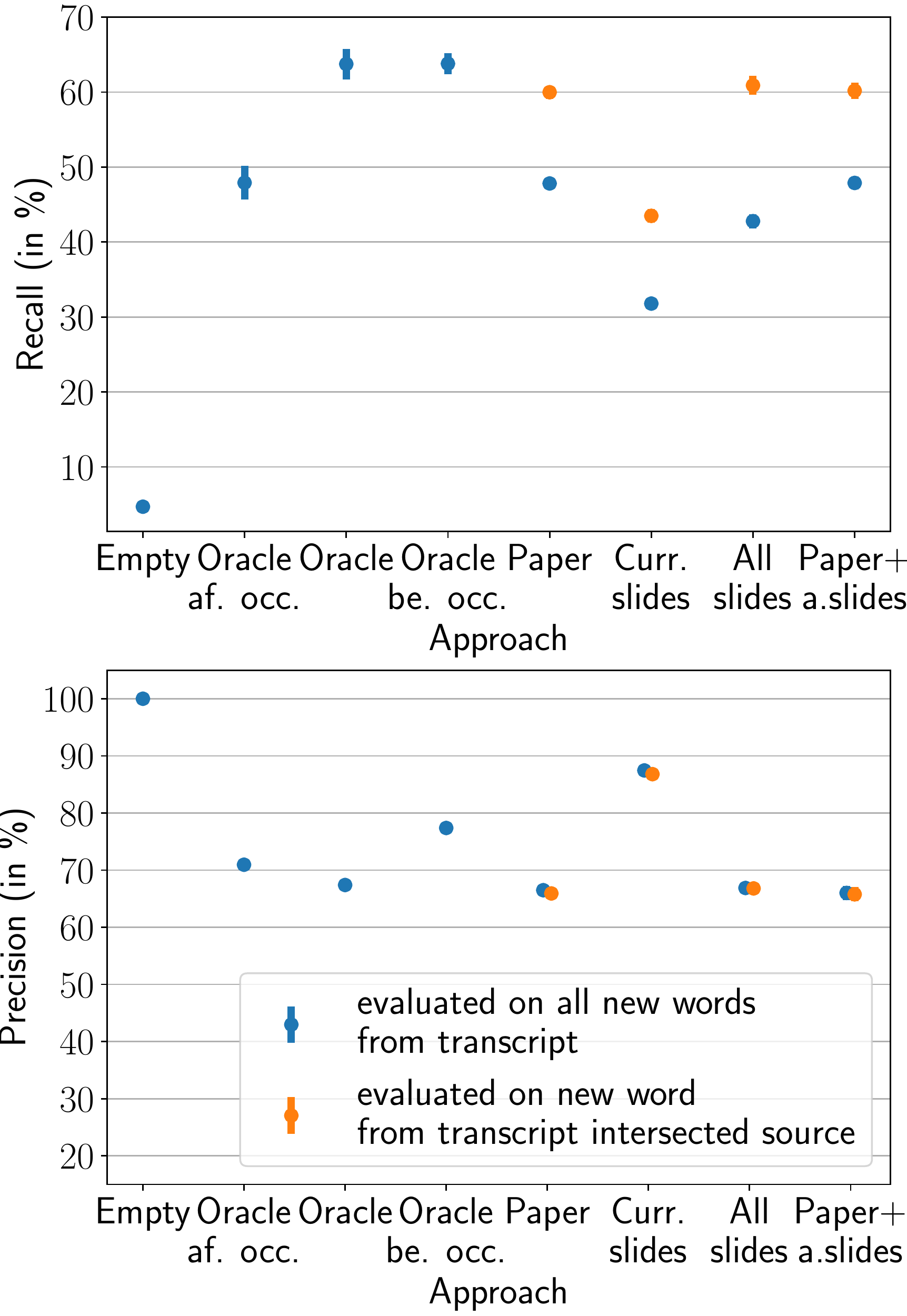}
    \caption{Recall and precision for the evaluation of the memory ASR-worker on the EMNLP testset.}
    \label{fig:testsetEMNLP}
\end{figure}

\begin{figure}[t]
    \centering
    \includegraphics[trim={18.1cm 0 0.25cm 0},clip,width=1.0\linewidth]{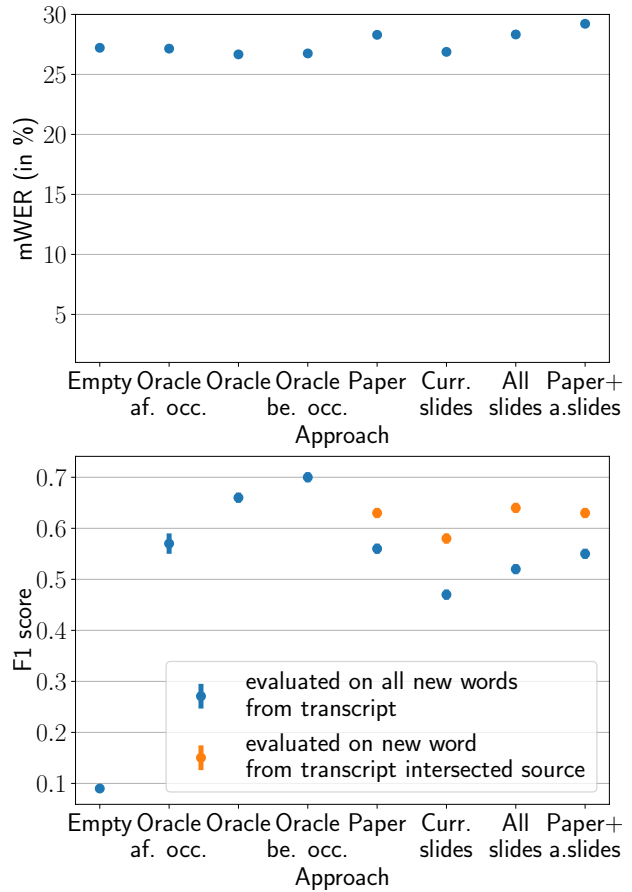}
    \caption{mWER and F1-score for the evaluation of the memory ASR-worker on the EMNLP testset.
    }
    \label{fig:testsetEMNLP2}
\end{figure}

In this scenario, we evaluate the system, when the list of new words is extracted from the reference transcript (Oracle). This simulates an operator introducing new misrecognized words in the memory. We use eight talks of the ELITR testset for the evaluation.

Dependent on the approach used (x axis in \cref{fig:testsetELITR,fig:testsetELITR2}), either an empty list of new words (Empty),
i.e. the baseline, 
or the full list of new words (Oracle) is used.
For the approaches ``* af. occ.'' and ``* be. occ.'', a new word is added to the list of new words \textit{after} the first \textit{occurrence} or \textit{before} the first \textit{occurrence} of that word, respectively. These two approaches simulate an operator correcting the output and either the corrected segment is reevaluated or not. The first condition reflects the case when the operator can quickly react to errors but cannot fix them once the segment has been shipped, the second condition represents the situation when either the shipping is delayed a little to give the operator a chance to introduce the correction, or when the overall system setup allows updating previous outputs. The second situation is common e.g. in re-translating systems such as ELITR \cite{bojar-etal-2021-elitr}.

Furthermore, we noticed, that sometimes false positives occurred, i.e. a word in the new words list is confused with a common word and outputted even though it is not in the audio and the reference transcript at that point. In the approaches marked with ``* ext. *'', we therefore \textit{extended} the list of new words and added these common words also to the new words list. This should help the model to distinguish between the common word and the new word.
These approaches simulate an operator adding common words to the memory when a false positive is observed.

The mWER segmenter \cite{matusov2005evaluating} is used to align the output segments with the reference segments. In \cref{fig:testsetELITR,fig:testsetELITR2}, we can see the results.
The recall, precision and F1 score is measured on the new words. The baseline model with empty memory performs poorly. For the other approaches, we see that it certainly helps to add the new word before the occurrence. The approaches with extended memory list produce substantially fewer false positives with the approach ``Oracle ext. be. occ.'' reaching an F1 score of $0.80\pm 0.02$.
Furthermore, we see that all approaches have similar word error rates (mWER) suggesting that the word error rate is not an appropriate measure for evaluating if important words are correctly recognized. Note that the talks are challenging and the transcript is not very clean and therefore the word error rates are relatively high.

We noticed that the performance of the ASR worker when used in online low-latency mode is not deterministic. This happens because the packets of audio are sent over the network and they can, dependent on the network latency, arrive earlier or later. Therefore, when the predetermined duration of audio is reached, the model can start processing a slightly smaller or longer audio input sequence.
Thus, as described above, the stable hypothesis found by the stability detection is not deterministic and therefore the same holds true for the ASR output. Therefore, we report mean and standard deviation performance over 16 runs as shown in \cref{fig:testsetELITR,fig:testsetELITR2}.

\subsection{Evaluation with Additional Sources}

In this scenario, we evaluate the system when the list of new words is extracted from additional sources. We conducted experiments on ten talks from the EMNLP 2020 conference. The list of new words was extracted from the paper or the slides (with optical character recognition), respectively, with the same method described above. For the approach ``Curr. slides'', the new words list was extracted from the previous, current and the upcoming slide.

In \cref{fig:testsetEMNLP,fig:testsetEMNLP2}, we see similar results for the approaches that were already evaluated on the ELITR testset (e.g. F1 scores of $0.66\pm 0.01$ vs. $0.65\pm 0.02$ for the oracle approaches).
For the approaches ``Paper'' and ``* slides'', we evaluate the performance on all the new words of the transcript and on the new words of the transcript intersected with the new words from the source. 
We differentiate between these two evaluation methods to show the effect of the new uttered word being actually available in the source or not.
The performance on the new words of the approach ``Paper'' is not much worse than the oracle approach (F1 scores of $0.66\pm 0.01$ vs. $0.56\pm 0.01$), especially when considering the evaluation only of new words present in the paper ($0.63\pm 0.01$), compared to an F1 score of $0.09\pm 0.00$ for the approach ``Empty''.

The performance when extracting the new words list from the slides is worse when evaluating on all new words from the transcript, possibly due to the used optical character recognition.
When evaluating only on new words present in the slides, the performance is better than the corresponding approach of extracting the new words from the paper. 
Combining the new words from the paper and the slides did not yield improvements.
The word error rate (mWER) stays almost the same for all approaches, however one can see that for the approaches with many words in the memory, the word error rate is a bit higher. We investigate this phenomenon further in section \ref{sec:add-exp}.

\subsection{Additional Experiments}
\label{sec:add-exp}

\begin{figure}[t]
    \centering
    \includegraphics[trim={0 0 0 0},clip,width=1.0\linewidth]{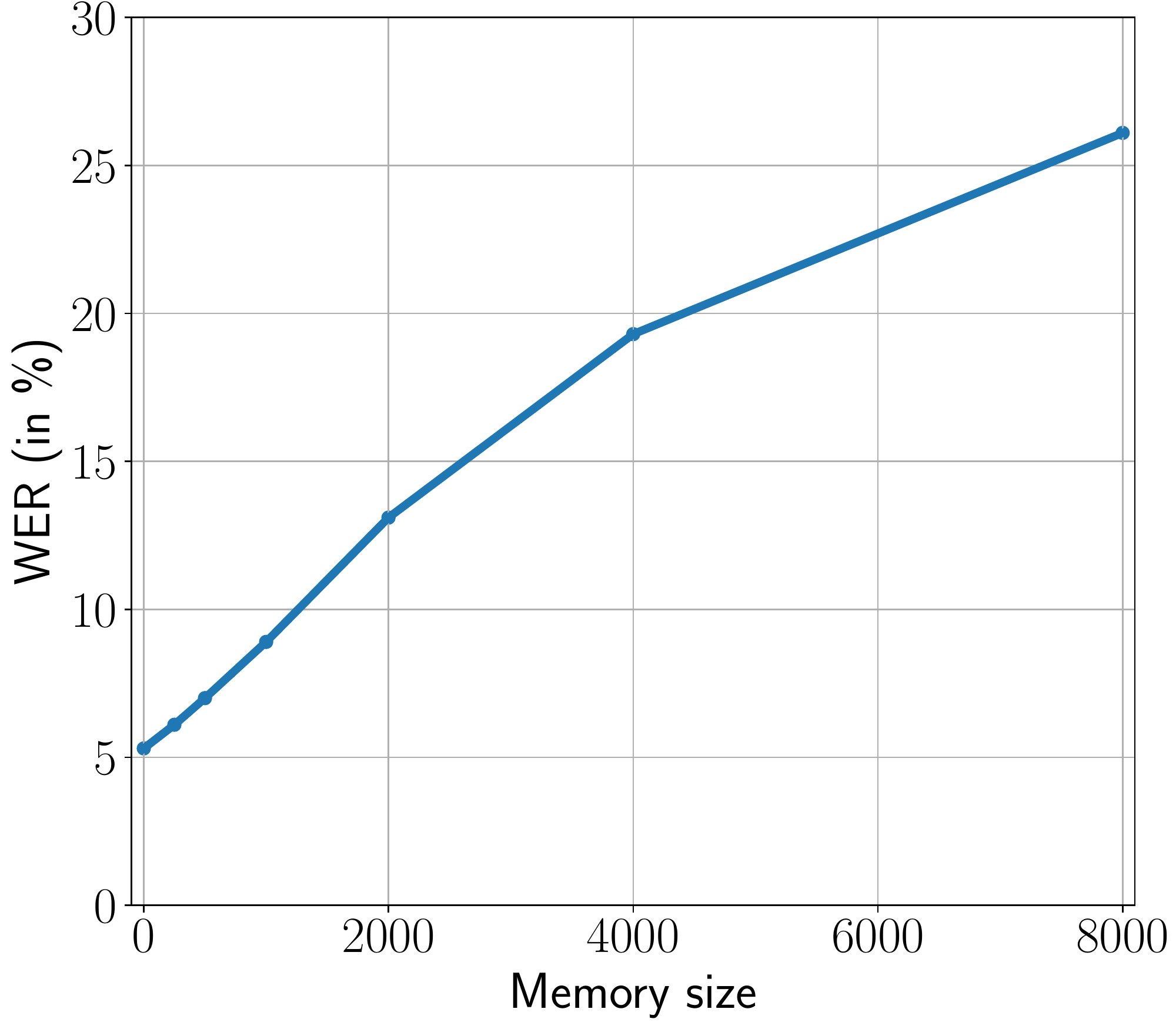}
    \caption{Evaluation of the memory ASR-worker on the tedlium testset with a large number of random words from the training dataset in the memory.}
    \label{fig:testsetMemorySize}
\end{figure}

We investigated the usage of a large memory and took a certain number of random words from the training dataset as memory. Then we decoded the tedlium testset \cite{hernandez2018ted}. The results can be seen in \cref{fig:testsetMemorySize}: a huge number of words in the memory result in a drastically worse word error rate. This happens since a lot of false positives are occurring.
Therefore, our approach is best used with a small number of new words the model should focus on.

As the final experiment, we examined the situation when a new word in the memory is not found (by the memory-attention layers). This can happen if the pronunciation of the word differs considerable from the ``common pronunciation'' as learned by the general model. To help the model recognize the new word anyway, we propose not to search for the new word but for the word the model outputs instead. So for example if we want to recognize ``his name is ron weasly" but the model would output ``his name is ron weesley" even if the word ``weasly" is in the memory, we would use the confused form ``weesley" in the memory-attention layer which searches through the memory and eventually use ``weasly" for memory-entry-attention layer which 
extracts information from 
from the memory.

Note that in this case, the scenario for the practical use is severely different. Instead of the information that a novel word will probably occur somewhere in the transcript, one now needs to have the information that at a specific point a word is wrong and another word is correct at that location.

We went through the false negatives of the new words testset from \cite{huber2021instant} and applied this approach.
As a result, we obtained that the accuracy on the new words testset increased from 90.4\% to 94.1\%.

\section{Conclusion}

We demonstrated an efficient method of acquiring a new words list given a source such as supplementary paper or slides and evaluated the trade-off between improving the recognition accuracy of new words and the occurrence of too many false positives in an online low-latency environment. 
We documented that standard WER does not reflect the success of recognition of these typically very important words.
We obtained an F1 score of up to $0.80$ evaluated on the recognition of new words.

\section{Acknowledgements}

We want to thank SlidesLive,\footnote{\url{https://library.slideslive.com/}} who provided us with the transcripts of the EMNLP talks.

The projects on which this paper is based were funded by the European Union under grant agreement No 825460 (ELITR), the Federal Ministry of Education and Research (BMBF) of Germany under the numbers 01IS18040A (OML) and 01EF1803B (RELATER) and the  Czech Science Foundation under the grant 19-26934X (NEUREM3).

\bibliographystyle{IEEEtran}

\bibliography{biblio}

\begin{thebibliography}{1}
\providecommand{\url}[1]{#1}
\csname url@samestyle\endcsname
\providecommand{\newblock}{\relax}
\providecommand{\bibinfo}[2]{#2}
\providecommand{\BIBentrySTDinterwordspacing}{\spaceskip=0pt\relax}
\providecommand{\BIBentryALTinterwordstretchfactor}{4}
\providecommand{\BIBentryALTinterwordspacing}{\spaceskip=\fontdimen2\font plus
\BIBentryALTinterwordstretchfactor\fontdimen3\font minus
  \fontdimen4\font\relax}
\providecommand{\BIBforeignlanguage}[2]{{%
\expandafter\ifx\csname l@#1\endcsname\relax
\typeout{** WARNING: IEEEtran.bst: No hyphenation pattern has been}%
\typeout{** loaded for the language `#1'. Using the pattern for}%
\typeout{** the default language instead.}%
\else
\language=\csname l@#1\endcsname
\fi
#2}}
\providecommand{\BIBdecl}{\relax}
\BIBdecl

\bibitem{huber2021instant}
C.~Huber, J.~Hussain, S.~St{\"u}ker, and A.~Waibel, ``Instant one-shot
  word-learning for context-specific neural sequence-to-sequence speech
  recognition,'' \emph{arXiv preprint arXiv:2107.02268}, 2021.

\bibitem{vaswani2017attention}
A.~Vaswani, N.~Shazeer, N.~Parmar, J.~Uszkoreit, L.~Jones, A.~N. Gomez,
  {\L}.~Kaiser, and I.~Polosukhin, ``Attention is all you need,'' in
  \emph{Advances in neural information processing systems}, 2017, pp.
  5998--6008.

\bibitem{pham2019very}
N.-Q. Pham, T.-S. Nguyen, J.~Niehues, M.~M{\"u}ller, and A.~Waibel, ``Very deep
  self-attention networks for end-to-end speech recognition,'' \emph{Proc.
  Interspeech 2019}, pp. 66--70, 2019.

\bibitem{nguyen2021superhuman}
T.-S. Nguyen, S.~Stueker, and A.~Waibel, ``Super-human performance in online
  low-latency recognition of conversational speech,'' 2021.

\bibitem{sltev:eacl:2021}
E.~Ansari, O.~Bojar, B.~Haddow, and M.~Mahmoudi, ``{SLTev: Comprehensive
  Evaluation of Spoken Language Translation},'' in \emph{Proc. of EACL Demo
  Papers}.\hskip 1em plus 0.5em minus 0.4em\relax Kyiv, Ukraine: ACL, 2021.

\bibitem{bojar-etal-2021-elitr}
\BIBentryALTinterwordspacing
O.~Bojar, D.~Mach{\'a}{\v{c}}ek, S.~Sagar, O.~Smr{\v{z}}, J.~Kratochv{\'\i}l,
  P.~Pol{\'a}k, E.~Ansari, M.~Mahmoudi, R.~Kumar, D.~Franceschini, C.~Canton,
  I.~Simonini, T.-S. Nguyen, F.~Schneider, S.~St{\"u}ker, A.~Waibel, B.~Haddow,
  R.~Sennrich, and P.~Williams, ``{ELITR} multilingual live subtitling: Demo
  and strategy,'' in \emph{Proc. of EACL System Demonstrations}.\hskip 1em plus
  0.5em minus 0.4em\relax ACL, 2021, pp. 271--277. [Online]. Available:
  \url{https://aclanthology.org/2021.eacl-demos.32}
\BIBentrySTDinterwordspacing

\bibitem{matusov2005evaluating}
E.~Matusov, G.~Leusch, O.~Bender, and H.~Ney, ``Evaluating machine translation
  output with automatic sentence segmentation,'' in \emph{Proceedings of the
  Second International Workshop on Spoken Language Translation}, 2005.

\bibitem{hernandez2018ted}
F.~Hernandez, V.~Nguyen, S.~Ghannay, N.~Tomashenko, and Y.~Est{\`e}ve,
  ``Ted-lium 3: twice as much data and corpus repartition for experiments on
  speaker adaptation,'' in \emph{International Conference on Speech and
  Computer}.\hskip 1em plus 0.5em minus 0.4em\relax Springer, 2018, pp.
  198--208.

\end{thebibliography}

\end{document}